\documentclass[conference]{IEEEtran}
\usepackage{blindtext, graphicx, caption}
\usepackage[table,xcdraw]{xcolor}
\usepackage{booktabs}
\usepackage{multirow}
\usepackage{tabularx}
\usepackage{url}

\usepackage[colorlinks = true,
            urlcolor  = blue]{hyperref}
\ifCLASSINFOpdf
\else
\fi

\begin{document}
%
\title{Benchmarking Scene Text Recognition in Devanagari, Telugu and  Malayalam\vspace{-1em}}


\author{
\IEEEauthorblockN{Minesh Mathew, Mohit Jain and C. V. Jawahar}
\IEEEauthorblockA{Center for Visual Information Technology, IIIT Hyderabad, India.}
\IEEEauthorblockA{ minesh.mathew@research.iiit.ac.in, mohit.jain@research.iiit.ac.in, jawahar@iiit.ac.in}
\vspace{0.7em}}



%


\maketitle
\thispagestyle{empty}

\begin{abstract}
Inspired by  the success of Deep Learning based approaches to English scene text recognition, we pose and benchmark scene text recognition for three Indic scripts - Devanagari, Telugu and Malayalam. Synthetic word images rendered from Unicode fonts  are used for training the recognition system. And the performance is bench-marked on a new  -\textit{ IIIT-ILST} dataset comprising of hundreds of real scene images containing text in the above mentioned scripts.  We use a segmentation free, hybrid but end-to-end trainable \textsc{cnn-rnn} deep neural network for transcribing the word images to the corresponding texts. The cropped word images need not be segmented into the sub-word units and the error is calculated and backpropagated  for the the given word image at once. The network is trained  using  \textsc{ctc} loss, which is proven quite effective for sequence-to-sequence transcription tasks. The \textsc{cnn} layers in the network learn to extract robust feature representations from word images. The sequence of features learnt by the convolutional block is transcribed to a sequence of labels by the \textsc{rnn+ctc} block.
The transcription is not bound by word length or a lexicon and is ideal for Indian languages which are highly inflectional.
\textsc{iiit-ilst} dataset, synthetic word images dataset and the script used to render synthetic images are available at \href {http://cvit.iiit.ac.in/research/projects/cvit-projects/iiit-ilst}{http://cvit.iiit.ac.in/research/projects/cvit-projects/iiit-ilst}.
\end{abstract}

\begin{IEEEkeywords}
Indian Languages, Scene Text, Indic Scripts, Synthetic data, CNN-RNN, Text Recognition, OCR 
\end{IEEEkeywords}

%
\IEEEpeerreviewmaketitle

\section{Introduction}
\subsection{Scene Text  Recognition}
The problem of scene text recognition deals with recognizing text in natural scene images. Traditionally text recognition was focused on recognizing printed text in documents. Such systems expected the images to be black and white, and in a document style layout comprising of text lines. The text in natural scenes in contrast appear in huge varieties in terms of layout, fonts and style. The traditional \textsc{ocr} systems do not generalize well to such a setting where inconsistent lighting, occlusion, background noise, and higher order distortions add to the problem complexity. Most works in this area, treat the scene text recognition problem as two sub-problems - detecting bounding boxes of words in an image and then recognizing the individual, cropped word images. Our work deals with the second problem, where a cropped word image need to be recognized.

Recognizing text appearing in natural scenes has become increasingly relevant today with the proliferation of mobile imaging devices. Text appearing in natural scenes  provide a great deal of information helpful in understanding what the whole image is about. To be able to recognize text in natural scenes will be quite useful in scenarios like autonomous navigation, assistive technologies for the visually impaired, traffic surveillance systems, mobile transliteration/translation technologies, mobile document scanners etc. Hence building a robust scene text recognition system will have a significant impact on many other problems involving computer vision.

Deep Learning based methods significantly improved the scene text recognition accuracies for English.
A Convolutional Neural Network (\textsc{cnn}) was used in ~\cite{wang, bissacco} to recognize individual characters in  a word image and predicted characters are combined for a word to output the transcription for the given word image. These methods required a good character segmentation algorithm to segment the word image into sub-word units. Such methods would not be suitable for Indic scripts where sub word segmentation is often difficult.
The problem was modelled as  image classification  in~\cite{jaderberg} where each word image was classified into word classes, drawn from a fixed size lexicon. This method, bounded by a lexicon is inherently not suitable for highly inflectional Indian languages. Another set of solutions learn common representations for word-label pairs and retrieval and recognition is performed on the learnt representations.
In~\cite{almazan}  word image and the text are embedded into a vector-subspace. This setting, enabled to model the scene text recognition problem as a retrieval problem - to retrieve the most satiable text  from the vector-subspace once a word image is given.

\begin{figure}[!]
\centering
\captionsetup{justification=centering}
\includegraphics[width=\columnwidth, height=150pt]{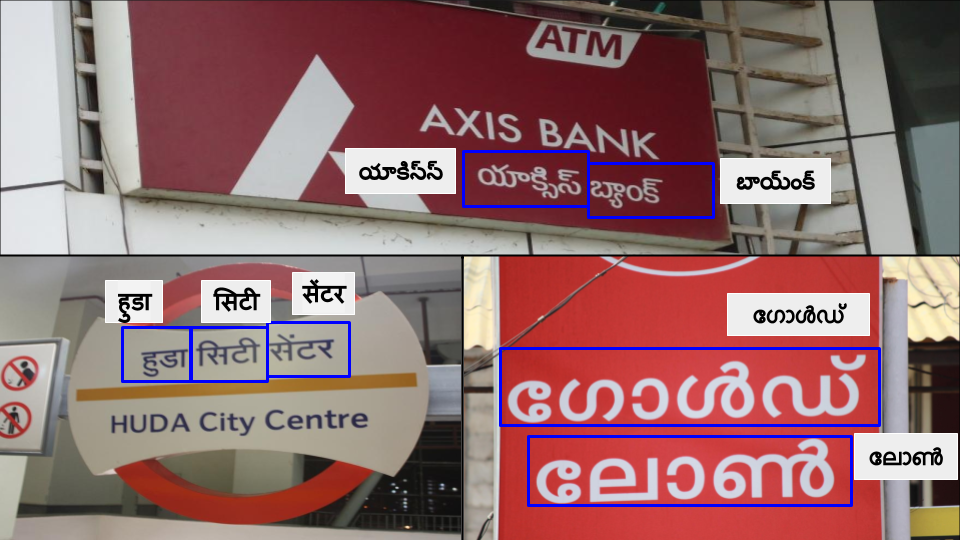}
\caption{Natural scene images, having text in Indic scripts; Telugu, Malayalam and Devanagari in the clock wise order }
\label{fig:fig6}
\end{figure}
The segmentation-free, transcription approach has been proven quite effective for Optical Character Recognition (\textsc{ocr}) in Indic Scripts~\cite{naveenicpr,mineshdas} and Arabic~\cite{hasan} where segmentation is often problematic.
Similar approach was followed for English scene text recognition in~\cite{su}. Handcrafted features derived from image gradients were used with a Bidirectional Long Short Term Memory (\textsc{lstm}) to map the sequence of features to a sequence of labels. Then  Connectionist Temporal Classification (\textsc{ctc})~\cite{graves06} loss was used  to computer the loss for the entire sequence of image features at once. Unlike the problem of \textsc{ocr}, scene text recognition required more robust features to yield results comparable to the transcription based solutions for \textsc{ocr}.
A novel approach combining the robust convolutional features and transcription abilities of \textsc{rnn} was introduced in~\cite{shi}. Here a 7 layer \textsc{cnn} stack is used at the head of an \textsc{rnn} + \textsc{ctc} transcription network. The network named as \textsc{crnn}  is end-to-end trainable and yielded better results than transcription using handcrafted features.
\subsection{Recognizing Indic Scripts}
Research in text recognition for Indic scripts has  mostly been centred around  the problem of printed text recognition; popularly known as \textsc{ocr}. Lack of annotated data, and inherent  complexities of the script and language were the major challenges faced by the community. Most of the early machine learning approaches which were effective for English \textsc{ocr}, were not easily adaptable to Indian languages setting for this reason. Even today, when modern machine learning methods could be used in a language/script agnostic manner, lack of annotated data remains as major challenge in case of Indian languages.
There had been few  works, which addressed the data scarcity  by using synthetic data. A synthetic dataset comprising of 28K word images was used in~\cite{pramodsynth} for training a nearest neighbour based Telugu \textsc{ocr}. The images were  rendered from a vocabulary of 1000 words, by varying font, font size, kerning and other rendering parameters.

Early attempts to recognize text in Indian scripts, often required a segmentation module which could segment word  into sub-word units like characters or \textit{akshara}s.
Lately the works in \textsc{ocr} started following segmentation-free approaches.  There have been works using  Hidden Markov Models (\textsc{hmm}s)~\cite{ocrhmm} and Recurrent Neural Networks (\textsc{rnn}s)~\cite{mineshdas}. Among these \textsc{rnn}s became quite popular choice for transcribing text words or lines directly into a sequence of class labels. \textsc{lstm} Networks used along with (\textsc{ctc}) loss enabled end-to-end training of a network which can transcribe from a sequence of image features to a sequence of characters.  This approach did not require segmenting words or line images into sub-units, and could handle variable length images and output label sequences. It has been proven quite effective for complex scripts like Indic Scripts and Arabic were segmentation of words into sub-word units is often difficult~\cite{naveenicpr,mineshdas,hasan}. Use of a Bidirectional LSTM  (\textsc{blstm})  enabled modelling past and future  contextual dependencies. This significantly helped in accurately predicting certain characters, particularly vowel modifiers (\textit{matras}) in Indic scripts. The challenges posed by Indic scripts and how the transcription approach helped to overcome those are discussed in detail in~\cite{mineshdas}. \textsc{ocr} using \textsc{rnn+ctc} has been using either raw pixels of the input image~\cite{hasan,mineshdas} or handcrafted features like profile based features~\cite{naveenicpr} as the input representation. We shall refer to such methods using \textsc{rnn+ctc} on handcrafted features as \textsc{rnn-ocr} hereafter.

\subsection{Scene Text Recognition for Indic Scripts}

Inspired form the success of \textsc{rnn-ocr}, we attempt to extend it to the problem of scene text recognition in Indian languages. Unlike English , there has been no works in scene text recognition in Indic scripts to the best of our knowledge.
Three major contributions of this paper can be listed down as

\begin{itemize}
  \item Synthetic scene text data - Word images are rendered using Unicode fonts for indic scripts, from a large vocabulary. The foreground and background texture, colors and other rendering parameters are varied to make the rendered images look similar to the real scene images. Fig.~\ref{fig:fig2} shows sample synthetic images rendered
  \item Real scene text data - A new, real scene text dataset - \textit{IIIT-ILST} was curated to benchamrk the recognition performance. This dataset comprises of hundreds of word images, for each script
  \item Benchmarking Scene text recognition - A segmentation free, end-to-end trainable system is employed. The hybrid \textsc{cnn-rnn} network used here, is trained purely on syntheticcally rendered word images. The trained network is then tested on the real scene text data.
\end{itemize}

The rest of the paper is organized as follows; Section II describes the hybrid \textsc{cnn-rnn} architecture we use. Section III presents  details of the rendering process used to generate the synthetic dataset, a brief summary of the \textit{IIIT-ILST} dataset and the network architecture used.
Quantitative and qualitative results, and a discussion on the results are presented in section IV
Section V concludes with the findings of our work.

\begin{figure}[!]
\centering
\captionsetup{justification=centering}
\includegraphics[width=\columnwidth]{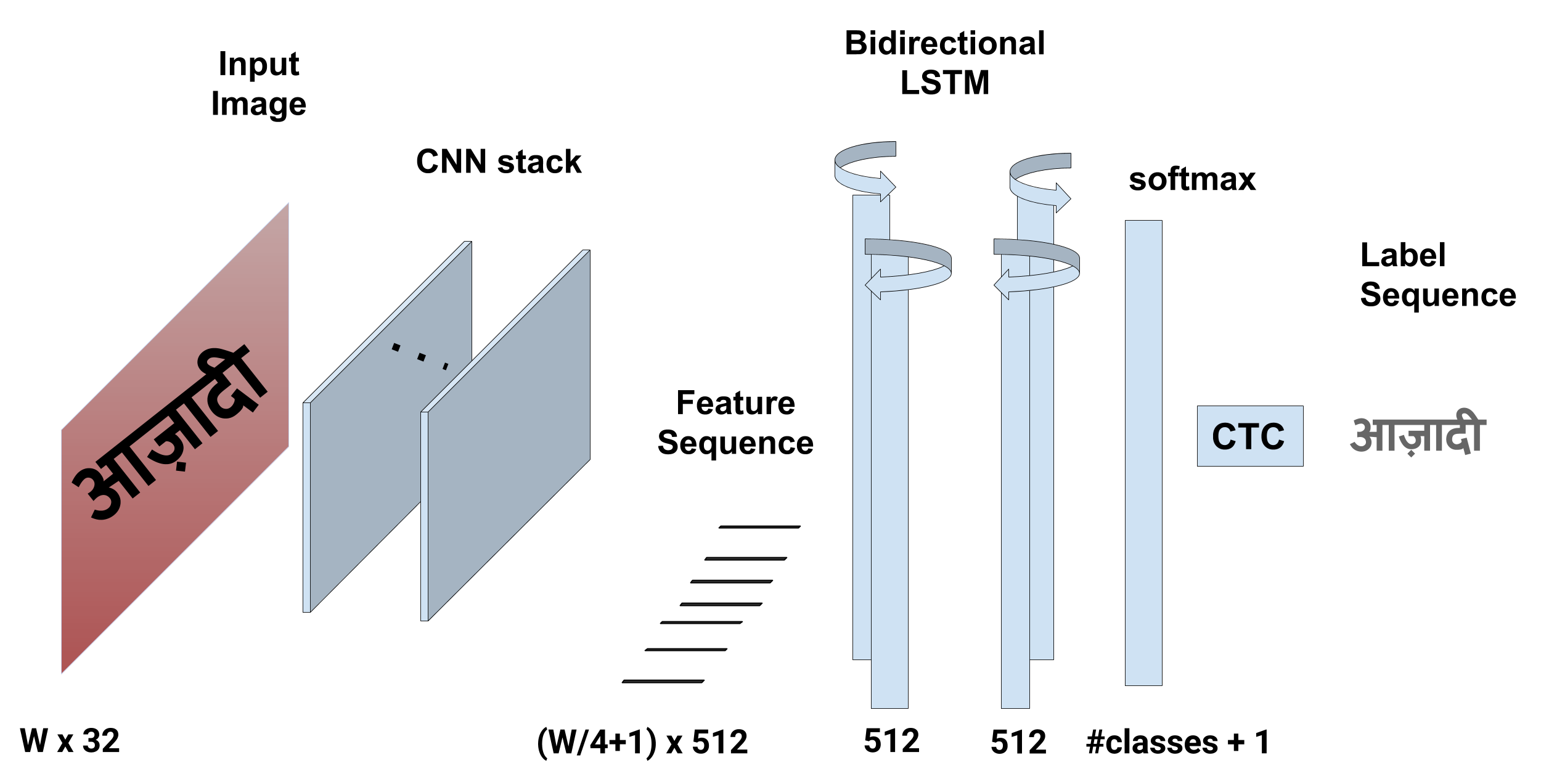}
\caption{Visualization of the Hybrid \textsc{cnn-rnn} network architecture used in this work.}
\label{fig:fig3}
\end{figure}

\section{\textsc{cnn-rnn} Hybrid Architecture for Transcription}
The hybrid \textsc{cnn-rnn} network consists of three components. The initial convolutional layers, middle recurrent layers and a final transcription layer. This network can be trained in an end-to-end fashion using the \textsc{ctc} loss and is not constrained by any language-specific lexicon. Consequently, any possible combination of the script's character-set can be recognized by the model.

The convolutional layers follow a \textsc{vgg}~\cite{simonyan} style architecture where the fully-connected layers have been removed. These layers obtain robust feature representations from the input images. The sequence of  features are then passed on to the recurrent layers which transcribe them into an output sequence of labels representing the  script's characters/glyphs. Transcription on to the output sequence of labels is performed using a \textsc{ctc} loss layer at the output.

All the images are scaled to a fixed height before being fed to the convolutional layers. The convolutional components then create a sequence of feature vectors from the feature maps by splitting them column-wise, which then act as inputs to the recurrent layers (Fig.~\ref{fig:fig3}). The convolutional features learnt by the network, are much more robust than the handcrafted features.
The layers of convolution, max-pool and element-wise activation functions are translation invariant as they operate on local regions. Hence, each column feature from the feature map corresponds to a rectangle region of the original image. These rectangular regions are themselves in the same order to their corresponding columns on the feature maps from left to right and hence can be considered as an image descriptor for that region. The recurrent layers following the convolutional layers, take each frame from these column-feature sequences and make predictions. 

The recurrent layers consist of deep \textsc{blstm} nets. 
Since the number of parameters of an \textsc{rnn} is independent of the length of its input sequence, \textsc{rnn}s are capable of handling variable length sequences. The network can  be unrolled as many times as the number of time-steps in the input sequence and hence, any sequence of characters/glyphs derived from the root script's character set can be predicted. 
This in our case helps to perform unconstrained recognition as the predicted output can be any sequence of labels derived from the entire label set.
Traditional \textsc{rnn} units (\textit{vanilla \textsc{rnn}s}) faced the problem of vanishing gradients~\cite{bengio} and hence \textsc{lstm} units are used which by design tackle the vanishing-gradients problem~\cite{ hochreiter} and remembers contextual dependencies for longer time.

For a typical text recognition problem, the transcription accuracy is benefitted if context from both directions (left-to-right and right-to-left) are available. A  \textsc{blstm}  is hence used to combine contexts from forward and backward orientations. Multiple such \textsc{blstm} layers can be stacked to make the network deeper and gain higher levels of abstractions over the image-sequences as shown in~\cite{graves13}.

The transcription layer at the top of the network is used to translate the predictions generated by the recurrent layers into label sequences for the target language. The \textsc{ctc} layer's conditional probability is used in the objective function as shown in~\cite{graves06}. The complete network configuration used for the experiments can be seen in Fig.~\ref{fig:fig3}.
The objective is to minimize the negative log-likelihood of conditional probability of ground truth. This objective function calculates a cost value directly from an image and its ground truth label sequence, eliminating the need of manually label all the individual components in sequence.

\section{Transcribing Scene Text in Indic Scripts}
Since scene text recogniton in Indic scripts has not been attempted before, we introduce two new datasets. A synthetic dataset, comprising of around 4 million images for each script and the \textit{IIIT-ILST} dataset comprising of real scene images for benchmarking the performance of the scene text recognition. The details are presented in the following two subsections.

\subsection{Synthetic Scene Text Dataset}
Synthetic data is now widely used within the computer vision community for problems where it is often difficult to acquire large amounts of training data. This approach to deal with the data scarcity issue has become popular with the advent of Deep Learning based methods, which are ever more data  hungry~\cite{mjsynth,ankush,virtualkitti}. The trend was prevalent in the area of text recognition, even before Deep learning based methods became popular. For low resource languages like Indian languages and Arabic, synthetically rendered text images have been in use for quite sometime~\cite{pramodsynth,apti,upti}. It has been widely accepted when~\cite{mjsynth} trained an English scene text recognition system trained purely on a synthetic dataset comprising of 8 million word images called \textit{MJSynth}. Since then all the works in English scene text recognition have been using this synthetic dataset alone as the training data. The use of such a huge corpus of word images along with state-of-the art deep learning methods yielded superior results in English scene text recognition. Recently, similar approach was followed in making  synthetic datasets for  scene text and video text recognition  for Arabic~\cite{mohitarabic}

\begin{figure}[!]
\centering
\captionsetup{justification=centering}
\includegraphics[width=\columnwidth, height=140pt]{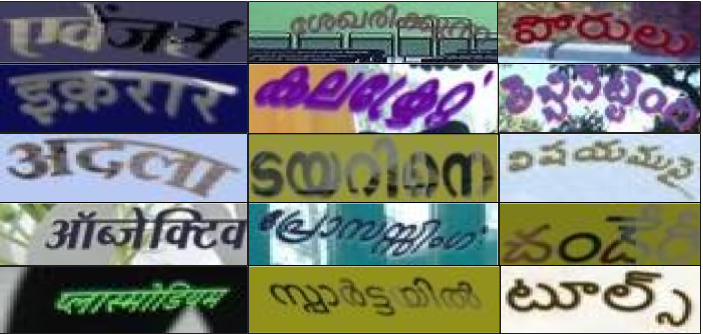}
\caption{Sample images for Hindi, Malayalam and Telugu from the synthetic dataset. }
\label{fig:fig2}
\end{figure}

In this work, the \textsc{cnn-rnn} hybrid architecture is purely trained on synthetically generated word images rendered from a large vocabulary - around 100K words for each script crawled from Wikipedia. The words are rendered into images using freely available Unicode fonts. Each word from the vocabulary is first rendered into the foreground layer of the image by varying the font, font size, stroke color, stroke thickness, kerning, skew and rotation along the horizontal line. Later a random perspective projective transformation is applied to the foreground image, which is then optionally blended with a random crop from a natural scene image. Finally the foreground layer is alpha composed with a background layer, which is either an image with uniform background color, or a random crop from a natural image. Details of the rendering process are presented here~\cite{syntharxiv}.

\begin{figure*}[htp]
\centering
\captionsetup{justification=centering}
\includegraphics[width=\textwidth, height=130pt]{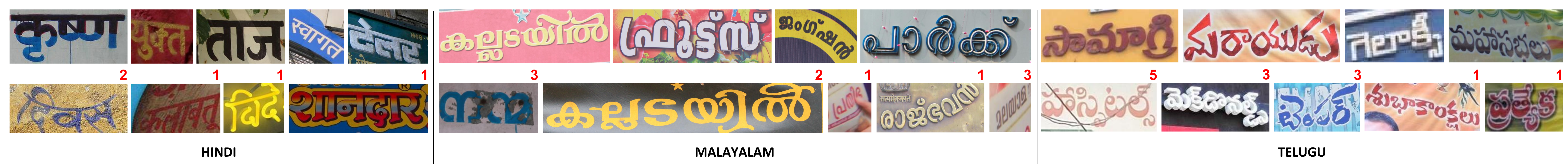}
\caption{Qualitative results of scene text  recognition for Indic scripts. For each script, the top-row shows images for which correct predictions were made while the bottom-row shows images with incorrect predictions. The number on top-right corner of each incorrectly predicted image is the \textit{Levenshtein distance} between its ground-truth and the prediction made by our model.}
\label{fig:fig4}
\end{figure*}

\subsection{IIIT-ILST Datasaet}
A new dataset for benchmarking the performance of scene text recogniton for Indic scripts was curated by capturing images with text in Indic scripts and compiling freely avaibale images from Google Images. The dataset consists of  hundreds of  scene images occuring in various scenrios like local markets, billboards, navigation and traffic signs, banners, graffiti. etc, and spans a large variety of naturally occuring image-noises and distortions. There are around 1000 scene text word images for each script extracted from these scene images. Each scene image is annotated by marking the bounding boxes of each word in the image, and typing the corresponding text in Unicode. Hence the dataset can also be used for scene text detection task. For the problem of scene text recognition addressed in this paper, we only deal with recognizing the individual word images. Hence in all the results discussed below, it is assumed that cropped word images are provided.

\subsection{Implementation Details}

The hybrid \textsc{cnn-rnn} network has it's  convolutional stack inspired  from the \textsc{vgg}-style architecture with minor modifications made to the layers to better fit a script recognition setting. In the 3rd and 4th max-pooling layers, the pooling windows used are rectangular instead of the usual square windows used in \textsc{vgg}. This helps us obtain wider features after the convolutions and hence more time-steps for the recurrent layers to unfold on. All input images are converted to grey scale and re-scaled to a fixed height of 32 pixels, while keeping the aspect ratio the same. The convolutional stack is followed by two \textsc{blstm} layers each of size 512. The second \textsc{blstm} layer is connected to a fully connected layer of size equivalent to the number of  labels + 1 (extra label for  \textit{blank}) . Lables in our case are Unicode points of the script for which the transcription is learnt. The label set include all the unique Unicode points found in the vocabulary used to render the synthetic dataset and basic punctuation symbols. Finally Softmax activation is applied to the outputs at the last year and the \textsc{ctc} loss is computed between the output probabilities and the expected, target label sequence.

To  accelerate the training process, two batch-normalization~\cite{ioffe} is performed after the 3\textsuperscript{rd} and 4\textsuperscript{th}  convolutional layers. We have observed that applying batch normalization after each convolutional layer yielded slightly poorer accuracies. To automate the process of setting optimization parameters, \textsc{adadelta} optimization~\cite{zeiler} is used while training the network using stochastic gradient descent (\textsc{sgd}). The transcription layers' error differentials are backpropagated with the forward-backward algorithm. While in the recurrent layers, the Backpropagation Through Time (\textsc{bptt})~\cite{werbos} algorithm is applied to calculate the error differentials.

\section{Results and Discussion}
We compare the transcription capabilities of our hybrid \textsc{cnn-rnn} network against an \textsc{rnn-ocr} style model.  \textsc{rnn-ocr}  could yield superior results for printed text recognition in Indic scripts~\cite{mineshdas}  even without a convolutional feature extraction block, since the printed text recognition is relatively an easier problem compared to the scene text recognition. We compare the performance of the hybrid architecture against \textsc{rnn-ocr} style approach, where raw image pixels are directly fed to the \textsc{rnn}. The \textsc{rnn-ocr} style approach used here is same as the one used for \textsc{ocr} in~\cite{mineshdas}.  
Table~\ref{tab-accuracies} presents a comparison of both the approaches. 

The performance has been evaluated using the following metrics; \textit{\textsc{crr} - Character Recognition Rate} and \textit{\textsc{wrr} - Word Recognition Rate}. In the below equations, \textit{RT} and \textit{GT} stand for recognized text and ground truth respectively.

{\fontsize{8}{4}\selectfont 
\[CRR = \frac{(nCharacters - \sum LevenshteinDistance(RT, GT))}{nCharacters}\]
\[WRR = \frac{nWordsCorrectlyRecognized}{nWords}\]
}

\begin{table}[!]
\centering
\captionsetup{justification=centering, margin=2em}
\caption{Performance Evaluation on \textit{IIIT-ILST} Dataset}
\label{tab-accuracies}
\begin{tabular}{|c|c|c|c|c|c|}
\hline
\cellcolor[HTML]{9B9B9B}& \multicolumn{5}{c|}{\cellcolor[HTML]{C0C0C0}Method}\\ \cline{2-6} 
\multirow{-2}{*}{\cellcolor[HTML]{9B9B9B}\textbf{IL SceneText}} & & \multicolumn{2}{c|}{\textsc{rnn-ocr}} & \multicolumn{2}{c|}{\textsc{hybrid cnn-rnn}} \\ \cline{1-1} \cline{3-6} 
\cellcolor[HTML]{C0C0C0}Script  & \multirow{-2}{*}{No. of Images} & \textsc{wrr (\%)} & \textsc{crr (\%)}& \textsc{wrr (\%)} & \textsc{crr (\%)} \\ \hline
Hindi   & 1150& 29.7   & 58.1  & 42.9 & 75.6 \\ \hline
Telugu  & 1211& 33.6   & 61  & 57.2 & 86.2 \\ \hline
Malayalam   & 807& 40.2   & 73.2  & 73.4 & 92.8 \\ \hline
\end{tabular}
\caption*{Comparing performance of the hybrid \textsc{cnn-rnn} model against the \textsc{rnn-ocr} style approach in~\cite{mineshdas} }
\end{table}

The tabular results further reinforce our hypothesis that addition of convolutional layers before the transcription block, improves the performance, since the convolutional layers output much robust representations than the raw pixel values of the input images or other handcrafted features.  A qualitative analysis of the trained models' transcription capabilities can be seen in Fig.~\ref{fig:fig4}. 


\begin{figure}[htp]
\centering
\captionsetup{justification=centering}
\includegraphics[width=\columnwidth]{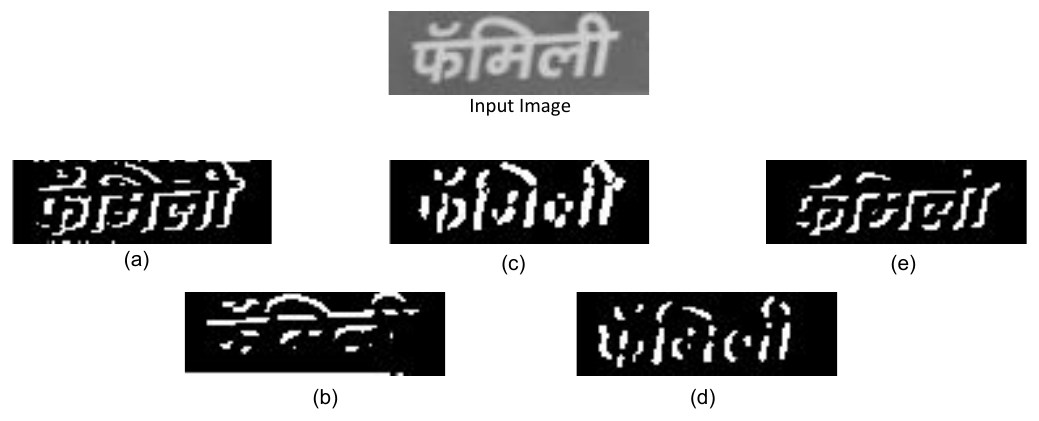}
\caption{\small{Visualization of activations learnt by the hybrid \textsc{cnn-rnn} model's convolutional layers. We observe that convolutional layers learn to detect edges and diacritics in the text and derive directional edge information from the source image. Notice how (a) detects all text edges while (b) focuses on \textit{matras} and \textit{top-connector-line} of Devanagiri script. Right-edges, left-edges and right-bottom edges are detected in (c), (d) and (e) respectively.}}
\label{fig:fig5}
\end{figure}

To gain further insights into the workings of the convolutional layers, we visualize the activations of these layers generated by performing a forward pass on the model using sample images~(Fig.~\ref{fig:fig5}). The model seems to be learning orientation-specific edge detectors in the initial convolutional layers. Additionally, the convolutional layers also grasp the ability to differentiate between main text body and the diacritics appearing in contiguation.

\section{Conclusion}
 We demonstrate that state-of-the-art deep learning based methods can be successfully adapted to some rather challenging tasks like scene text recognition in Indic scripts. The newer script and language agnostic approaches are well suited for low resource languages like Indian languages where the traditional methods often involved language specific modules. The success of \textsc{rnn}s in sequence learning problems has been instrumental in the recent advances in speech recognition and image to text transcription problems. This came as a boon for Indic scripts where segmentation of words into sub word units are often troublesome. The sequence learning approach could directly transcribe the images and also model the context in both forward and backward directions. With better feature representations and learning algorithms available, we believe the focus should now shift to harder problems like scene text recognition. We hope the introduction of a new dataset and the initial results would instill an interest among the researchers to pursue this field of research further. In future we hope to increase the size of the real scene dataset and to add images in other indic scripts too. With a reasonably big real scene text corpus, some percentage of the images shall be used to fine-tune the network which now is purely trained on synthetic images. We believe that the fine-tuning would considerably improve the performance.

\textbf{Acknowledgment.} Minesh Mathew is supported by TCS Research Scholar Fellowship


\ifCLASSOPTIONcaptionsoff
  \newpage
\fi



%

\end{document}